\documentclass[runningheads]{llncs}

\usepackage[T1]{fontenc}
\usepackage{graphicx}
\usepackage{tcolorbox}
\usepackage[disable]{todonotes}
\usepackage[skip=2pt]{caption}
\usepackage{subcaption}
\usepackage{pgfplots}
\pgfplotsset{compat=1.17}
\usepackage{booktabs}
\usepackage[numbers,sort]{natbib}
\usepackage{amsmath}
\usepackage[symbol]{footmisc}
\usepackage{xcolor}
\usepackage{acronym}
\usepackage[inline]{enumitem}
\usepackage{tikz}
\usetikzlibrary{positioning, shapes.geometric, arrows.meta, calc, math}
\usepackage{hyperref}
\hypersetup{
  colorlinks   = true,    
  urlcolor     = mydarkblue,    
  linkcolor    = mydarkblue,    
  citecolor    = mydarkpink      
}

\usepackage{contour}
\usepackage{ulem}

\usepackage{sansmath}
\AtBeginEnvironment{tikzpicture}{\sansmath}
\AtEndEnvironment{tikzpicture}{\unsansmath}

\contourlength{0.8pt}
\newcommand{\ul}[1]{%
  \uline{\phantom{#1}}%
  \llap{\contour{white}{#1}}%
}

\newcommand{\mpara}[1]{\smallskip\noindent{\it #1}}

\newcommand{\cmdr}{\textsc{Command-R$+$}}

\acrodef{GTR}{generate-then-retrieve}
\acrodef{IR}{information retrieval}
\acrodef{LLM}{large language model}
\acrodef{NLI}{natural language inference}
\acrodef{NLP}{natural language processing}
\acrodef{RAG}{retrieval-augmented generation}
\acrodef{RTG}{retrieve-then-generate}

\colorlet{mypink}{blue!20!red!30!white}
\colorlet{mygreen}{green!60!blue!40!white}
\colorlet{myred}{blue!10!red!50}
\colorlet{mydarkred}{blue!20!red!80}
\colorlet{mydarkestred}{blue!30!red!70!black}
\colorlet{mydarkpink}{red!60!blue!60}
\colorlet{mydarkestgreen}{green!60!blue!80!black}
\colorlet{mydarkgreen}{green!60!blue!90}
\colorlet{myblue}{red!10!blue!30!white}
\colorlet{mydarkblue}{red!20!blue!70!white}
\colorlet{mylightgrey}{black!5}

\tikzset{%
    myarrow/.style={-Stealth, thick},
    process/.style={rectangle, draw, thick, align=center, minimum height=0.7cm, minimum width=1.2cm},
    processsmal/.style={rectangle, draw, thick, align=center, minimum height=0.7cm, minimum width=1.2cm,font=\footnotesize},
    data/.style={cylinder, shape border rotate=90, draw, thick, minimum height=0.7cm,
    minimum width=1cm, align=center, aspect=0.3},
}%

\begin{document}
\title{Correctness is not Faithfulness\\ in RAG Attributions}
%
%
\author{Jonas Wallat$^{*,}$\inst{1}
\and
Maria Heuss$^{*,}$\inst{2}
\and
Maarten de Rijke\inst{2}
\and
Avishek Anand\inst{3}
}

\def\thefootnote{*}\footnotetext{Equal contribution}
%
\authorrunning{J. Wallat et al.}
%
\institute{L3S Research Center Hannover, Germany 
\\\email{jonas.wallat@l3s.de} \and
University of Amsterdam, The Netherlands \\
\email{<m.c.heuss|m.derijke>@uva.nl}
\and
TU Delft, The Netherlands \\ \email{avishek.anand@tudelf.nl}
}

\maketitle
\begin{abstract}
Retrieving relevant context is a common approach to reduce hallucinations and enhance answer reliability. Explicitly citing source documents allows users to verify generated responses and increases trust. Prior work largely evaluates \textbf{citation correctness} -- whether cited documents support the corresponding statements. But citation correctness alone is insufficient. To establish trust in attributed answers, we must examine both \textbf{citation correctness} and \textbf{citation faithfulness}. In this work, we first disentangle the notions of citation correctness and faithfulness, which have been applied inconsistently in previous studies. Faithfulness ensures that the model’s reliance on cited documents is genuine, reflecting actual reference use rather than superficial alignment with prior beliefs, which we call \textit{post-rationalization}. We design an experiment that reveals the prevalent issue of post-rationalization, which undermines reliable attribution and may result in misplaced trust. Our findings suggest that current attributed answers often lack citation faithfulness (up to $57\%$ of the citations), highlighting the need to evaluate correctness and faithfulness for trustworthy attribution in language models.

\end{abstract}
\section{Introduction}

\begin{figure}[ht]
    \centering
        {\input{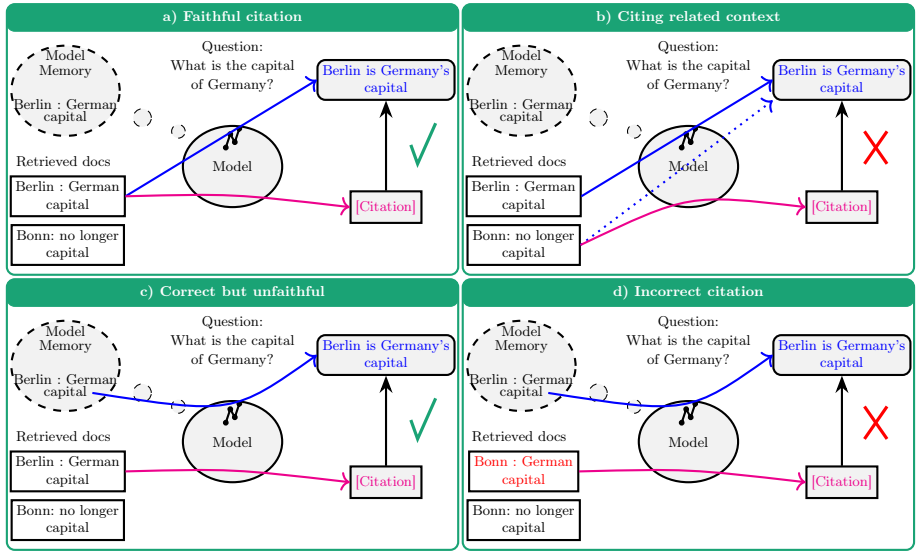}}
    \caption{Different answer scenarios for the query ``What is the capital of Germany?'' (a)~The ideal case, i.e., a correct citation that is faithful to the answer's generation process. 
    (c)~A correct but unfaithful citation, where the model post-rationalizes a citation to fit its prior. 
    (b)~A citation referring to the context that was used during the answer generation but does not contain the statement itself. 
    (d)~An incorrect citation.}
    \label{fig:intro_faithful_vs_correct}
\end{figure}

Trustworthiness of \ac{IR} systems is key to their usage in high-stakes scenarios~\citep{jacovi-2021-formalizing,high-level-2019-ethics}.
Hallucinations occur when \acp{LLM} generate plausible but incorrect or fabricated information, hindering the trustworthiness of IR systems that employ LLMs~\cite{DBLP:journals/corr/abs-2311-01463}. 
One method to address the challenge posed by hallucinations in text output is to enable generated text that is directly grounded in retrieved source documents and accompanied by citations~\cite{rashkin2023measuring,bohnet2023attributed}. While citations do not necessarily prevent hallucinations, they increase the verifiability of generated claims~\cite{liu2023evaluating}. 
Grounded text generation~\cite{gao2020robust:gtg} has been studied in \ac{NLP} tasks like summarization and question answering, aiming to produce content derived from specific sources -- either provided by users or, more recently, retrieved through mechanisms like \ac{RAG}~\cite{lewis2020retrieval}. 
Attributed text generation, often implemented as RAG systems, focuses on generating text with supporting evidence, ensuring coherence, contextual relevance, and grounding in verifiable sources by generating text alongside supporting evidence~\cite{bohnet2023attributed}.
The current evaluation of attributed text focuses on the overall \textbf{correctness of the answer} and the \textbf{correctness of citations}, which is based on the agreement between attributed statements and the information that can be found in the referenced source documents. 

Citation correctness, sometimes called answer faithfulness, e.g.~\citet{gao2023retrieval}, 
measures the extent to which cited documents support a generated statement. 
We argue that ensuring correction is not always enough.
For tasks like legal information retrieval~\cite{maxwell2008concept} or medical question answering~\cite{lee2006beyond}, where the retrieved documents are complex, and the answers are sensitive to model biases, simple fact-checking or correctness evaluation might prove difficult and require a nuanced understanding of the documents themselves. 
Yet, in those domains, unwarranted trust in generated answers can have detrimental consequences.
We need to understand the model's reasoning process to verify that the model correctly used the cited documents to produce its answer and that it was not answered from its parametric memory through \textbf{post-rationalization}, i.e., where models may cite sources to fit preconceived notions rather than genuine retrieval. 
We use the term \textbf{citation faithfulness} to describe whether the citation accurately depicts the model's reasoning process. It is hard to determine citation faithfulness if we only evaluate the correctness of citations. 
Figure~\ref{fig:intro_faithful_vs_correct} illustrates the differences between faithful and unfaithful behavior as well as correct and incorrect citations.  

When building trustworthy IR systems that offer self-explanations -- in this case, citations -- we should strive to convey the system's decisions accurately. 
Only if the produced citations are faithful to the underlying processes can we enable justified trust (as opposed to misplaced trust if faithfulness breaks down).

Our contributions in this work are threefold. First, we offer coherent notions of attribution and citation in the context of grounded generation and introduce the concept of citation faithfulness. Second, we propose desiderata for citations that go beyond correctness and accuracy and are needed for trustworthy and usable systems. Third, we emphasize the need to evaluate the faithfulness of citations by studying post-rationalization.
Our experiments reveal the existence of unfaithful behavior, with up to 57\% of citations being post-rationalized.

Our work on disentangling citation correctness and faithfulness in grounded text generation using LLMs aims to create more reliable IR systems by ensuring accurate and contextually faithful citations. By focusing on post-rationalization, we enhance accountability, helping IR systems avoid propagating biases or misinformation, thus promoting ethical standards in information dissemination\footnote{Code available at \url{https://github.com/jwallat/RAG-attributions}}.
\section{Related Work}

We summarize relevant background and position our work w.r.t.\ the evaluation of attributed generation, faithfulness in interpretability, and faithfulness of self-explanations.
The area of knowledge conflicts \cite{DBLP:journals/corr/abs-2403-08319} aims to understand information flow and whether answers originate from parametric memory or contextual information \cite{DBLP:conf/acl/NeemanAHCSA23,DBLP:journals/corr/abs-2310-00935}. 
Its goal of understanding models is similar to ours but it has a different focus (full answers vs.\ citations) and  is therefore out of scope.

\subsection{LLMs and Attributions}\label{sec:RW-LLMS}

\begin{figure}[t]
    \centering
        {


\begin{tikzpicture}[scale=0.76, transform shape]
    \tikzmath{
    \boxwidth = 3.9;
    \boxheight = 6;
    \spacing = 0.1;
    }
     \draw[local bounding box=rect0, rounded corners=3pt, draw=mydarkgreen,thick]  (0,0) rectangle ($(0,0) + (\boxwidth,\boxheight)$);
        \begin{scope}[
            x={($(rect0.south east)-(rect0.south west)$)},
            y={($(rect0.north west)-(rect0.south west)$)},
            shift={(rect0.south west) - (2,2)}
        ]
            \draw[rounded corners=3pt,
            draw=mydarkgreen,
            fill=mydarkgreen] (0, 0.9) rectangle (1.0,1.0);
            \node at (0.5, 0.95) {\textbf{\color{white}{GTR (Post-hoc Attr.)}}};
            \node (question) at (0.18, 0.8) {Question};
            \node[data,draw=mydarkpink] (data2) at (0.85, 0.4) {DB};
            \node[process, draw=mydarkblue] (llm) at (0.5, 0.6) {LLM};
            \node (answer) at (0.15, 0.4) {Answer};
            \node (attributed_answer) at (0.5, 0.05) {\ul{Attributed Answer}};
            \draw[myarrow] (question.south) -- (llm.north west);
            \draw[myarrow] (llm.south west) -- (answer.north);
            \draw[myarrow] (data2.south west) -- (attributed_answer.north);
            \draw[myarrow] (answer.east) -- (data2.west)
                node[midway,above, sloped] {find}
                node[midway,below, sloped] {support};
            \draw[myarrow] (answer.south) -- (attributed_answer.north);
        \end{scope}
    \draw[local bounding box=rect, rounded corners=3pt, draw=mydarkgreen,thick]  ($($(rect0.south east)$) + (\spacing,0)$) rectangle
    ($($($(rect0.south east)$) + (\spacing,0)$) + ($(\boxwidth,\boxheight)$)$);
        \begin{scope}[
            x={($(rect.south east)-(rect.south west)$)},
            y={($(rect.north west)-(rect.south west)$)},
            shift={(rect.south west)}
        ]
            \draw[rounded corners=3pt,
            draw=mydarkgreen,
            fill=mydarkgreen] (0, 0.9) rectangle (1.0,1.0);
            \node at (0.5, 0.95) {\textbf{\color{white}{RTG (Post-hoc Attr.)}}};
            \node (question2) at (0.18, 0.8) {Question};
            \node[data,draw=mydarkpink] (data21) at (0.85, 0.8) {DB};
            \node[data,draw=mydarkpink] (data22) at (0.85, 0.4) {DB};
            \node[process, draw=mydarkblue] (llm2) at (0.5, 0.6) {LLM};
            \node (answer2) at (0.15, 0.4) {Answer};
            \node (attributed_answer2) at (0.5, 0.05) {\ul{Attributed Answer}};
            \draw[myarrow] (question2.south) -- (llm2.north west);
            \draw[myarrow] (question2.east) -- (data21.west);
            \draw[myarrow] (data21.south west) -- (llm2.north east);
            \draw[myarrow] (llm2.south west) -- (answer2.north);
            \draw[myarrow] (data22.south west) -- (attributed_answer2.north);
            \draw[myarrow] (answer2.east) -- (data22.west)
                node[midway,above, sloped] {find}
                node[midway,below, sloped] {support};
            \draw[myarrow] (answer2.south) -- (attributed_answer2.north);
        \end{scope}
    \draw[local bounding box=rect2, rounded corners=3pt, draw=mydarkgreen,thick]  ($($(rect.south east)$) + (\spacing,0)$)  rectangle     ($($($(rect.south east)$) + (\spacing,0)$) + ($(\boxwidth,\boxheight)$)$);
        \begin{scope}[
            x={($(rect2.south east)-(rect2.south west)$)},
            y={($(rect2.north west)-(rect2.south west)$)},
            shift={(rect2.south west)}
        ]
            \draw[rounded corners=3pt,
            draw=mydarkgreen,
            fill=mydarkgreen] (0, 0.9) rectangle (1.0,1.0);
            \node at (0.5, 0.95) {\textbf{\color{white}{RTG (direct Attr.)}}};
            \node (question3) at (0.18, 0.8) {Question};
            \node[data,draw=mydarkpink] (data31) at (0.85, 0.8) {DB};
            \node[process, draw=mydarkblue] (llm3) at (0.5, 0.6) {LLM};
            \node (attributed_answer3) at (0.5, 0.05) {\ul{Attributed Answer}};
            \draw[myarrow] (question3.south) -- (llm3.north west);
            \draw[myarrow] (question3.east) -- (data31.west);
            \draw[myarrow] (data31.south west) -- (llm3.north east);
            \draw[myarrow] (llm3.south) -- (attributed_answer3.north)
                node[midway,right, align=left] {Prompt:\\please cite\\the sources};
        \end{scope}
    \draw[local bounding box=rect3, rounded corners=3pt, draw=mydarkgreen,thick]  ($($(rect2.south east)$) + (\spacing,0)$)  rectangle ($($($(rect2.south east)$) + (\spacing,0)$) + ($(\boxwidth,\boxheight)$)$);
        \begin{scope}[
            x={($(rect3.south east)-(rect3.south west)$)},
            y={($(rect3.north west)-(rect3.south west)$)},
            shift={(rect3.south west)}
        ]
            \draw[rounded corners=3pt,draw=mydarkgreen,fill=mydarkgreen] (0, 0.9) rectangle (1.0,1.0);
            \node at (0.5, 0.95) {\textbf{\color{white}{Verifiable} RTG}};
            \node (question4) at (0.18, 0.8) {Question};
            \node[data,draw=mydarkpink] (data41) at (0.85, 0.8) {DB};
            \node[process,draw=mydarkblue] (llm4) at (0.5, 0.6) {LLM};
            \node (attributed_answer4) at (0.5, 0.05) {\ul{Attributed Answer}};
            \node (c0) at (0.5, 0.35) {\LARGE \textbf{?}};
            \draw[myarrow] (question4.south) -- (llm4.north west);
            \draw[myarrow] (question4.east) -- (data41.west);
            \draw[myarrow] (data41.south west) -- (llm4.north east);
            \draw[myarrow] (llm4.south) -- (c0.north);
            \draw[myarrow] (c0.south) -- (attributed_answer4.north);
        \end{scope}
\end{tikzpicture}
}
    \caption{Different methods of attribution generation regarding their likelihood for un-faithful behavior and post-rationalization (with approaches more likely of faithful behavior on the right).}
    \label{fig:grounded_gen_approaches}
\end{figure}
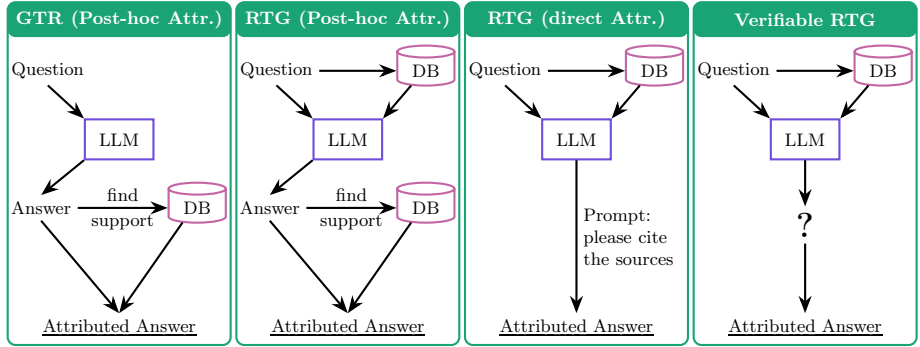

Supplying LLM-generated answers with attributions aims to improve the quality of the generated answers \cite{gao2023enabling}, reduce hallucination \cite{tonmoy:2024:arxiv:hallucinationsurvey}, and improve users' trust \cite{menick:2022:arxiv:gophercite} in the generated outputs. 
Methods for generating attributed answers range from prompting \cite{gao2023enabling}, adding post-hoc attributions \cite{gao2023enabling,thoppilan:2022:arxiv:lamdba}, and training paradigms \cite{menick:2022:arxiv:gophercite,thoppilan:2022:arxiv:lamdba,DBLP:journals/corr/abs-2305-14908,DBLP:conf/iclr/AsaiWWSH24,DBLP:conf/naacl/Ye0AP24} to generation-planning for more fine-grained citations \cite{slobodkin2024attribute}. 
Figure~\ref{fig:grounded_gen_approaches} provides an overview of common methods.
The simplest method is \ac{GTR}, a paradigm in which a model produces an answer (without attributions), and supporting evidence is added in a subsequent step \cite{gao2023enabling,bohnet2023attributed}. 
\Ac{RTG} operates similarly, but the model produces the (unattributed) answer after seeing both the question and the retrieved documents. 
As with GTR, RTG produces attributions in a second retrieval step, independent of the initially retrieved documents \cite{DBLP:conf/naacl/Ye0AP24}. 
Thus, both GTR and RTG have post-hoc attributions, which are unfaithful to the model by design, i.e., the citation does not reflect the model's decision-making during the answer generation process. 
It is, however, possible to directly generate attributed answers by prompting the RTG model to do so \cite{bohnet2023attributed,gao2023enabling}. The resulting attributed answer \textit{may} be faithful to the model's decision process, but we lack guarantees. 
As we show below, there is a significant chance of unfaithful behavior.
The ultimate goal of attributed answer methodologies is to verify that certain information in the answer \textit{originates} from the source document. 

\subsection{Evaluation of Attributed Generation}
\label{sec:RW-Evaluation}
Attributed generation is a complex process that requires evaluation across multiple dimensions. One dimension is the \textit{usability} of the generated response, which includes factors like fluency and perceived utility~\cite{liu2023evaluating}. Traditionally, these factors have been assessed through user studies and automatic evaluation methods~\cite{gao2023enabling}.
Other important dimensions include \textit{answer relevance}, which measures how well the response addresses the question, and context relevance, which looks at the compactness of the retrieved context~\cite{es2023ragas}. 
Datasets like HAGRID~\cite{kamalloo:2023:arxiv:hagridrageval} are useful for evaluation, with human evaluations of the informativeness and attributability of the responses, which can be used to measure overlap with gold citations~\cite{djeddal2024evaluation}. 
\citet{weller2024according} use the QUIP-Score, a method based on n-gram comparisons, to measure grounding and quoting from model pre-training data.

Next to the generated answer, the citation to the referenced document needs to be evaluated, too.  
To this end, prior work often uses \ac{NLI} classifiers~\cite{gao2023rarr,bohnet2023attributed}. 
These help evaluate citation precision, which measures the average correctness of citations, and comprehensiveness/citation recall, which quantifies the proportion of accurately cited statements in all statements~\cite{djeddal2024evaluation,li2023survey}.
The correctness of citations is a major focus in prior work~\cite{zhang2024towards,djeddal2024evaluation,rashkin2023measuring,saad2024ares,adlakha2024evaluating,li2023survey,roychowdhury2024evaluation,mayfield:2024:sigir:Eval_ML_Reports,gao2023retrieval}. 
We differentiate between citation correctness and the related but distinct aspect of \textbf{citation faithfulness}. Citation faithfulness requires a causal relationship between the cited document and the generated statement, an area that has so far received little attention.

\subsection{Faithfulness in Interpretability}\label{sec:RW-interpretability}
\todo[inline]{Misplaced trust if we don't have faithfulness https://linnk.ai/insight/nlp-ai/faithfulness-vs-plausibility-in-explanations-from-large-language-models-LoCRbYLO/}

In \ac{RAG} attributions, (citation) faithfulness has not been studied much. In contrast, the evaluation of faithfulness of explanations has been studied extensively. 
Here, faithfulness refers to how accurately an explanation reflects the model's decision-making process, clearly differentiating it from explanation plausibility~\cite{jacovi2020towards}. 
It lacks a universally accepted formal definition and is often defined in an ad-hoc manner~\citep{lyu2024towards}.
Faithfulness establishes a causal relationship. 
Various methods have been proposed for evaluating faithfulness:
\begin{enumerate*}[label=(\roman*)]
\item axiomatic evaluation, 
\item predictive power evaluation, 
\item robustness evaluation, 
\item perturbation-based evaluation, 
\item white-box evaluation, and 
\item human perception evaluation~\cite{lyu2024towards}.
\end{enumerate*}
Twelve desirable properties of explanations have been identified by~\citet{nauta2023anecdotal}, including correctness (of explanations), which is equated with faithfulness. 
Overall, the concepts of faithfulness and correctness appear entangled in the explainability literature. 
We take a step towards disentangling those two aspects for attributed text. 
Inspired by \citet{lyu2024towards}, we consider the causal relationship between the attributed text and generated answer to be a fundamental condition of faithful attribution.

\subsection{Faithfulness of LLM self-explanations}\label{sec:RW-self-explanations}
Self-explanations are explanations that an LLM is prompted to generate along with the answer to a posed question. Self-explanations have been divided into 
\begin{enumerate*}[label=(\roman*)]
\item chain-of-thought (CoT) reasoning, which involves generating a sequence of intermediate steps that lead to the response~\cite{wei2022chain}, 
\item token importance, which highlights tokens that significantly influence the response generation~\cite{li2015visualizing,wu2020perturbed}, and 
\item counterfactual explanations, which provide insights into
how different inputs might lead to a different response~\cite{agarwal2024faithfulness}. 
\end{enumerate*}
Faithfulness of self-explanations has recently received attention~\cite{agarwal2024faithfulness,turpin2024language,lyu2023faithful,lanham2023measuring}, with work on evaluating faithfulness~\cite{turpin2024language,lanham2023measuring} and its importance in contrast to plausibility~\cite{agarwal2024faithfulness}. 
There is high variation in how
much LLMs use CoT on different tasks, some relying upon it heavily, others merely generating it in a post-hoc manner~\citep{lanham2023measuring}. 
We view attributed generation that generates citations along with the text, rather than post-hoc, as a special class of self-explanation. 
We use a similar evaluation strategy as was previously used for the evaluation of faithfulness for CoT explanations~\cite{turpin2024language} to show that similar faithfulness concerns arise for attributed generation as for CoT reasoning. 
We identify the problem of post-rationalization, which is closely related to post-hoc reasoning~\cite{lanham2023measuring}.

\section{Attributions}

In the context of attributed text generation in \ac{RAG}, an answer may contain references to documents emphasizing that certain information originates from the referenced document. Merriam-Webster defines the verb \textit{to attribute} as explaining (something) by indicating a \textit{cause}, emphasizing the causal nature.\footnote{\url{https://www.merriam-webster.com/dictionary/attribute\#h2}} 

\subsection{Notation}\label{sec:attributions-notations}
Let $A=\{a_i\}_i$ be a set of retrieved documents and let $s$ be a factual statement that needs to be grounded in the retrieved documents $A$. 
A citation  $cit: s\mapsto a_j\in A$, or simply $(s,a_j)$, connects a statement to a document that supports the stated statement. 
We use the term \textit{attribution} to refer to the referenced document $a_j$ or the process of referencing source documents. 

\begin{tcolorbox}[colframe=mydarkgreen,title=Example 1: Attributed Answer]
\textbf{Question}: Whats the biggest penguin in the world? \\
\textbf{Answer}: The \ul{Emperor Penguin}~[0] is the \ul{tallest} [0] or biggest penguin in the world.
\end{tcolorbox}

\noindent%
In Example 1, ``tallest'' would be a statement $s$ attributed to document $a = 0$ through the citation $(\text{"tallest"}, 0)$. We note that many attributed statements are underspecified. Therefore, we distinguish between the statement (``tallest'') and the underlying \textit{claim} (``Emperror penguin: tallest: in the world''). When attribution generation is integrated with answer generation, citations can be considered a special form of self-explanation, other forms being chain-of-though explanations~\cite{wei2022chain}, explain-then-predict and predict-then-explain frameworks~\cite{camburu2018snli}, and counterfactuals~\cite{chen2024do}. 

\subsection{Desiderata for Good Attributions}\label{sec:attributions-desiderata}

Several dimensions can make attribution good or bad. We discuss the \textit{correctness}, \textit{faithfulness}, \textit{appropriateness}, and \textit{comprehensiveness} of a citation (Table~\ref{tab:desiderata}).

\begin{table}[h]
\small
\caption{Desiderata for good attributions.}
\label{tab:desiderata}
\centering
\resizebox{\textwidth}{!}{%
\begin{tabular}{ll}
\toprule
\textbf{Desideratum} & \textbf{Description} \\
\midrule
Correctness & Attribution accurately represents the content of the cited document \\
Appropriateness & Attribution is relevant and meaningful, not noisy or irrelevant \\
Comprehensiveness & Attribution covers all the key points in the answer \\
Faithfulness & Attribution reflects the actual process leading to the answer \\
\bottomrule
\end{tabular}}
\end{table}


\smallskip\noindent%
\textbf{Correctness.}
Most importantly, good citations should be correct, meaning that the cited documents should support the generated statement. Ensuring correctness in attribution is crucial for maintaining the integrity and reliability of the information being presented. However, there are several ways in which the outputs of an LLM can be right or wrong. 

\mpara{Hallucinated attributions.}
Attributions that do not exist, i.e., when a model hallucinates a reference to a non-existing document, are relatively easy to spot.
LLMs without a retrieval component, such as the early versions of ChatGPT, especially, commonly generate broken links or hallucinate titles and authors of the source document from which certain information should come.

\mpara{Wrong answers.} 
A direct way in which an LLM-generated answer can be wrong is if the statement itself is wrong, not matching the ground truth answer. 
This is the property that is evaluated most frequently in the open-domain QA and attribution literature~\citep[e.g.,][]{lewis2020retrieval,djeddal2024evaluation,bohnet2023attributed}. Wrong answers can result from hallucinations or correct attributions from a document containing false information. Hence, the generated answer is incorrect in this case, yet wrong answers may include correct citations. 

\mpara{Wrong citations.}
Attributions can be incorrect, for example, by misrepresenting the content of the attributed documents or by attributing claims from document $a$ to document $b$. In these cases, the citation $(s,b)$ is incorrect. 
Compared to answer correctness, less work focuses on the correctness of attributions. Attributions are usually evaluated by testing if the attributed document implies the statement. To do so, recent work employs NLI models \cite{bohnet2023attributed,gao2023rarr,djeddal2024evaluation}.   

\smallskip\noindent%
\textbf{Appropriateness \& Comprehnsiveness -- \textit{What} do we cite?}
Besides unfaithful behavior and incorrect attributions, bad citations may (appear to) be inappropriate or non-comprehensive and, therefore, dilute our understanding or evaluation of the answer. Appropriateness of attributions means that the attribution should be relevant, understandable, and meaningful; comprehensiveness refers to covering all the key points in the answer. The question of how much we need to cite and whether attributions cover the important claims are less prominent in current evaluations, but these aspects may heavily skew the results. 

\begin{tcolorbox}[title=Example 2: Inappropriate Citations,colframe=mydarkgreen]
\textbf{Question}: how long was gabby in a coma in the choice \\
\textbf{Answer}: In the \ul{novel} [0, 4] \ul{the choice} [0, 3, 4], Gabby is in a coma for three months.
\end{tcolorbox}

\mpara{Inappropriate citations.}
In Example 2, neither citation offers much value given the question. Attributing the title ``the choice'' provided in the question to documents 0, 3, and 4 offers no additional insights. On the contrary, when evaluating the quality of the provided citations, common approaches average over all existing citations. A large number of such \textit{low-value} citations, which re-state information from the question, may heavily skew the evaluation metrics. 

\mpara{Short statements -- What is the actual claim?} Capturing a comprehensive, standalone statement in an LLM-generated response that maintains its specificity even when separated from the rest of the text can be a complex task. The statement is often reduced to a single word or concept, subtly referring to other parts of the generated response. In our example, it remains ambiguous as to what the highlighted word ``novel'' pertains to (i.e., the actual claim). This lack of clarity makes interpreting and evaluating such references more challenging.

\mpara{For which statements do we need a citation?}
An answer may contain several citations, but one may be missing for the factual answer to the question. In the above example, the focus of the question is the time that Gabby spent in a coma (``three months''). This is the most critical statement in the answer and should be attributed to a source document. The above answer is not comprehensive since a central requested fact is not attributed to any source. 

\smallskip\noindent%
\textbf{Faithfulness -- Right for the \textit{wrong} reason?}
Can an attribution be correct and still be bad? 
Like model explanations, attributions can be right for the wrong reason.
To judge whether an attribution is right for the wrong reason, it is key to understand the internal model processes and understand whether a document $a$ 
was considered during answer generation. 
If $a$ is cited for another reason, then the attribution is not faithful to the underlying model behavior. 
Importantly, unfaithful attributions might still be factually correct and, therefore, difficult to spot -- yet unfaithful attributions foster misguided trust. 

\mpara{Post-rationalization.}
We hypothesize that post-rationalized attributions are a special case of unfaithful behavior. In this setting, an LLM's parametric memory produces an answer to the question, and the model looks for support in the documents in some shallow way (e.g., by token-matching). The resulting citation is not faithful since the attribution maps to a document, not to the model's internal knowledge. Let us consider Example 3:  
\begin{tcolorbox}[title=Example 3: The Faithfulness Post-rationalization Correctness Trilemma,colframe=mydarkgreen]
\textbf{Question}: What is the capital of Germany? \\
\textbf{Answer}: The capital of Germany is \ul{Berlin} [1, 2]\\

\textbf{Document 1}: The capital of Germany is Berlin [...] \\
\textbf{Document 2}: Berlin has the best night-life [...]\\

\textbf{Faithful (right for the right reason)}: Citing document 1 because the LLM used document 1's information to generate the answer\\
\textbf{Post-rationalized} but \textbf{correct}(right for the wrong reason): Citing document 1 because the model knows the answer and finds a document that agrees with its priors\\
\textbf{Post-rationalized} and \textbf{wrong}: Citing document 2 because the model knows the answer, and the answer token is mentioned in document 2.
\end{tcolorbox}
\noindent
Since the outputs in both the faithful and unfaithful cases are identical (citing document 1), unfaithful behavior is hard to identify. We conjecture that for the evaluation of faithfulness, we need both the \textit{attributions} and the \textit{process} by which they have been generated.

\section{Citation Faithfulness}
\label{sec:faithfulness}

The Cambridge Dictionary defines \textit{faithful} as ``true or not changing any of the details, facts, style, etc. of the original.''\footnote{\url{https://dictionary.cambridge.org/us/dictionary/english/faithful}} The explainability literature states that ``faithful explanation should accurately reflect the reasoning process behind the model’s prediction''~\cite{jacovi2020towards}. 
\citet{lyu2024towards} state that faithfulness establishes causality, referring to an example where ``what is known by the model'' does not necessarily correspond to ``what is used by the model in making predictions.''

Prior work on attributed answer generation defines \textit{answer faithfulness} as the extent to which the cited document supports the generated statement~\cite{zhang2024towards}. 
Answer faithfulness considers the answer itself rather than the citation. In the context of the citation, this property is often called the correctness of the citation. In this work, we define \textbf{citation faithfulness} and disentangle the concepts of \textit{answer faithfulness}/\textit{correctness} and \textit{citation faithfulness}. Prior work on attributed answers often has defined faithfulness loosely, for example, as ``whether the selected documents influence the LLM during the generation''~\cite{DBLP:journals/corr/abs-2406-13663}. We take inspiration from the rich literature on the faithfulness of explanations and define the faithfulness of citations through a casual dependency of the generated answer and the referenced document.

\subsection{Towards a Definition of Citation Faithfulness}
We offer the following definition of citation faithfulness.

\begin{tcolorbox}[title=Definition of Citation Faithfulness,colframe=mydarkgreen]
Let $s$ be a generated statement. Let $A = \{a_i\}_i$ be a set of documents that the model has retrieved as context. We call $(s,a_j)$ a faithful citation if: 
\begin{itemize}
    \item $a_j \in A$, 
    \item $s$ is supported by $a_j$ (correctness), and
    \item $s$ is causally impacted by $a_j$. 
\end{itemize}
\end{tcolorbox}

\noindent%
The second condition, often referred to as the ``correctness'' of the citation, has been a focal point in previous studies evaluating RAG attribution. Correctness usually tests whether a statement is supported by the attributed document (measured by NLI models). However, while correctness is a necessary condition for faithfulness, it is insufficient. For a citation to be deemed faithful, the model must also rely causally on the cited document to generate the answer. The evaluation of this causal dependence of the model output on the cited statement has been largely overlooked, which is why we advocate for increased attention for the topic in future research.

We recognize that our definition of faithfulness is somewhat abstract. 
As \citet{lyu2024towards} note, formulating a concrete definition with a single sufficient test for evaluating the faithfulness of an explanation is challenging. 
Therefore, a set of more tangible necessary conditions with corresponding tests should be established in practice. 
These can assist in approximating the level of faithfulness of specific explanations.
Consider the following examples of more concrete necessary conditions for faithful attribution.
Assuming that a citation $(s,a)$ is faithful, the following should hold: 
\begin{enumerate}[label=(\arabic*)]
    \item If the relevant information in the cited document $a$ is altered, the model should either provide a different generated statement $s$ or modify the decision-making process. 
    This could involve using different evidence $a'$ or the model's memory to generate the answer.
    \item Adding irrelevant documents to the context should not affect the attribution, provided that the answer remains unchanged.
\end{enumerate}
In Section~\ref{sec:Experiments}, we design an experiment that tests this second necessary condition and shows evidence for the phenomenon of post-rationalization. 

\section{Post-Rationalization -- A Study of Unfaithful Behavior}\label{sec:Experiments}
We study attributions of a prominent RAG model and produce evidence of unfaithful behavior.  
Evaluating faithfulness requires carefully designed tests and an understanding of internal model processes~\cite{jacovi2020towards}. This is to ensure that the attribution matches the underlying decision processes.    

\mpara{Aim.} We investigate a particular case of unfaithful behavior, post-rationalization, i.e., the process in which a model comes up with a prior answer without regard to the documents and then searches retrieved documents to find supporting evidence. The following experiment establishes the presence of unfaithful behavior in existing attributions by \ac{RTG} models.

\mpara{Setup.}
Cohere's \cmdr{} model is a ``RAG-optimized'' LLM specifically trained to produce grounded answers.\footnote{\url{https://cohere.com/blog/command-r-plus-microsoft-azure}} It has 104B parameters and a context length of 128k tokens, which we use in 4-bit quantization to run on a single NVIDIA A100 GPU. We evaluate \cmdr{}'s attributions on the NaturalQuestions QA dataset, containing 1,444 real user questions answered by Wikipedia pages \cite{kwiatkowski:2019:TACL:NaturalQuestions}. We use the temporally-aligned KILT \cite{kilt_tasks} Wikipedia dump\footnote{Available here: \url{https://huggingface.co/datasets/facebook/kilt\_tasks}.} as a retrieval base. Following~\cite{cuconasu:2024:arxiv:powerofnoise}, we split passages into chunks of 100 tokens and prepend the title of the page to the chunks. 
We index the resulting chunks and, for each query, retrieve the top 30 documents using BM25. 
We rerank the 30 retrieved documents using ColBERT v2 \cite{santhanam:2022:nacl:colbertv2} and feed the top 5 documents together with the question into \cmdr{}. 

We use the grounded generation prompt template provided by Cohere.\footnote{\url{https://huggingface.co/CohereForAI/c4ai-command-r-plus}} The grounded generation pipeline with \cmdr{} follows four steps: 
\begin{enumerate*}[label=(\roman*)]
\item predict the relevance of the retrieved documents;
\item prediction which documents should be cited;
\item produce an answer without citations, and 
\item one with citations. 
\end{enumerate*}
This setup makes \cmdr{} a \acf{RTG} model with direct attributions via prompting (see Figure~\ref{fig:grounded_gen_approaches}). 
We selected an instance from this class of models since its chances of faithful behavior are higher than in the case of post-hoc attributions.

\mpara{Experiments.}
\label{sec:post-rationalization-exp}
We devise the following experiments to better understand the extent to which \cmdr{} post-rationalizes citations.
We aim to investigate whether the model performs token matching for its citations, so we 
\begin{enumerate*}[label=(\roman*)]
\item generate attributed answers for QA pairs, and 
\item select statements from these answers and append them to other documents. 
Since statements are, on average, 2--4 tokens, they mostly contain short concepts such as ``Emperor penguin'' or ``The Choice,'' which should not be cited when appearing without factual context. 
We append these adversarial statements into three kinds: random documents, documents predicted to be relevant but never cited, and documents cited for other information in the attributed answer. The created dataset with adversarial documents consists of 1344 QA pairs (random), 702 (relevant but not cited), and 829 (cited for other reasons). In step
\item we again generate attributed answers, but this time with our adversarial documents. In the case that the adversarial document was created from a random document, we append it to the list of documents in the context. If the original document was part of the context, we substitute it with the adversarial one. Lastly, 
\item we observe whether the model now cites our adversarial documents for the statements selected in step (ii)). We operate under the assumption that citing documents that just randomly contain the statement (``Emperor penguin'') indicates \textit{post-rationalization}.
\end{enumerate*}

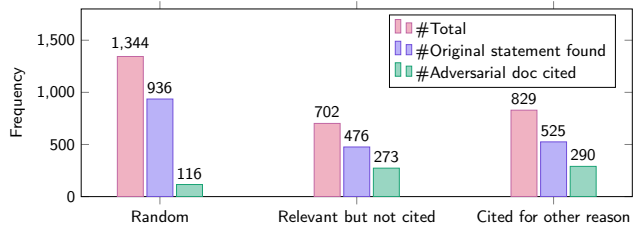
\begin{figure}[t]
    \centering
    \scalebox{.69}{
    
    
    

\begin{tikzpicture}[font=\sffamily]
    \begin{axis}[
        ybar,
        bar width=.5cm,
        width=\textwidth,
        height=5.2cm,
        enlarge x limits=0.2,
        legend style={at={(0.5,-0.15)},
          anchor=north,legend columns=1,
            legend cell align=left},
        legend pos= north east,
        symbolic x coords={Random, Relevant but not cited, Cited for other reason},
        xtick=data,
        ylabel={Frequency},
        nodes near coords,
        ymin=0,
        ymax=1800,
    ]
    \addplot[fill=mypink,draw=mydarkpink] coordinates {(Random,1344) (Relevant but not cited,702) (Cited for other reason,829)};
    \addplot[fill=myblue,draw=mydarkblue] coordinates {(Random,936) (Relevant but not cited,476) (Cited for other reason,525)};
    \addplot[fill=mygreen,draw=mydarkgreen] coordinates {(Random,116) (Relevant but not cited,273) (Cited for other reason,290)};
    \legend{\#Total, \#Original statement found, \#Adversarial doc cited}
    \end{axis}
\end{tikzpicture}}
    \caption{Results of the post-rationalization tests. We measure the cases in which the model cited our adversarial document (which had the previously cited statement appended). Since we also change the input, the model is not guaranteed to produce the same statements again. Therefore, we also include the number of cases where we could match the old statement.}
    \label{fig:post_ration_exp}
\end{figure}

\mpara{Results.}
The results are presented in Figure~\ref{fig:post_ration_exp}.
First and foremost, we note that recovering the old statement in the newly generated answer worked in 63--70\% of the cases. This is necessary to understand if the adversarial document has been cited for the same statement. By injecting the statement into random documents and passing them to the model, the model cited these documents in 12\% (116/936) cases.
Interestingly, the number of adversarial documents cited is much higher when forging relevant but uncited documents (57\%) and documents cited for different reasons (55\%). 
Based on our results, we conjecture post-rationalization to be a \textit{common phenomenon}. 

\section{Discussion and Outlook}
Our results are the first step toward understanding unfaithful behavior in RAG systems due to post-rationalization. We focus on attributions, where a \textit{faithful} attribution should signify the origin of the corresponding information.  
If the parametric model memory is used to generate an answer, a faithful model should either cite itself or just omit a citation. 
Several issues necessitate a principled approach to measuring faithfulness in future research. 
A primary obstacle in automatically detecting unfaithful behavior is developing robust evaluation frameworks for detection procedures. 
The challenges we encountered are reminiscent of those seen in explainability research within IR and other fields, where ensuring validity in attribution metrics remains difficult \cite{camburu2019itrustexplainerverifying,lyu2023faithful,DBLP:conf/emnlp/BastingsEZSF22}. 
A lack of ground truth, as well as the inherently interpretative nature of attribution for RAG systems, presents a challenge for constructing evaluation criteria that can accurately identify unfaithful outputs.

We suggest using evaluation strategies from explainability in IR, such as deliberate data contamination techniques \cite{singh2020model,idahl2021towards}, model probing to gain first insights into specific model capabilities \cite{sen:2020:sigir:probing,wallat2023probing,formal:2022:ecir:matchyourwords,wallat2024causal}, or reverse engineering parts of decision process \cite{chen:2024:arxiv:reverserelevance}.  
However, validating LLM-based attributions introduces new challenges that call for developing novel evaluation paradigms. We have proposed a preliminary test designed to assess faithfulness. 
This test, however, implicitly assumes that the model internals, or in other words, where the model looks and based on what it generates the answer, do not change through the insertion of additional irrelevant documents. To verify this assumption, an investigation of the model's internal states during answer generation would be necessary. 
Subsequent work could apply recent findings in understanding internal model processes to the problem of faithful attributions.

Our work underscores the importance of establishing control settings that yield conclusive evidence regarding faithfulness in model-generated content. We advocate for rigorous counterfactual setups to establish a more reliable foundation for attribution-based evaluations. Such setups could provide a framework to better understand the causal relationships in model behavior and confirm the validity of attribution methods within controlled environments.

Finally, our study provides the first step towards the human-verifiability of LLM-generated attributed text, offering an essential resource for end-users. 
Our research shows that LLM-generated content cannot be taken as faithful by default. In this regard, our work contributes valuable considerations for human reviewers tasked with evaluating and verifying LLM outputs, thereby supporting informed and trustworthy interactions with AI-generated text.
\section{Conclusion}
We have tackled the problem of faithful attribution for answer generation with LLMs. With the goal of verifiability in mind, we have formalized notions of attribution and citation faithfulness, clearly differentiating them from citation correctness. 
We provide empirical evidence of unfaithful citation behavior through post-rationalization in Command-R+, a state-of-the-art LLM trained for the RAG task, by measuring the impact of short text insertions into irrelevant documents on the generated citations. 
While prior work has focused mostly on evaluating the correctness of the generated citations, we argue that citation faithfulness is necessary for trustworthy RAG systems in high-stakes decision-making and decision-support. 
We acknowledge that this work's empirical analysis is limited. Future research will be necessary to establish the existence of post-rationalization and unfaithful behavior for a broader set of models and data selections. 
We advocate for more research in this area, especially on evaluating citation faithfulness. In the interpretability literature, this challenge is known to be difficult to solve due to the lack of ground truth. 
Additionally, further research into the impact of attribution on user trust could provide greater insight into the significance of this issue.




%
%
%
\bibliographystyle{splncs04nat}
\bibliography{references.bib}
%




\end{document}